\documentclass[letterpaper, 10 pt, conference]{ieeeconf}  %

\IEEEoverridecommandlockouts                              %

\usepackage[bookmarks=true]{hyperref}
\usepackage{cite}
\usepackage{amsmath,amssymb,amsfonts}
\usepackage{algorithmic}
\usepackage{graphicx}
\usepackage{textcomp}
\usepackage{xcolor}
\usepackage{placeins}
\usepackage{comment}
\usepackage[capitalise]{cleveref} %
\usepackage{wrapfig}
\usepackage{placeins}
\usepackage[caption=false, font=footnotesize]{subfig}
\usepackage{multirow}

\title{\LARGE \bf
Comparison of Optimization-Based Methods for Energy-Optimal Quadrotor Motion Planning
}

\author{Welf Rehberg, Joaquim Ortiz-Haro, Marc Toussaint and Wolfgang Hönig%
\thanks{The authors are with TU Berlin, Germany. \{w.rehberg, joaquim.ortizdeharo, toussaint, hoenig\}@tu-berlin.de}
\thanks{The research was funded by the Deutsche Forschungsgemeinschaft (DFG, German Research Foundation) - 448549715 and by the German-Israeli Foundation for Scientific Research (GIF) grant I-1491-407.6/2019. Joaquim Ortiz-Haro thanks the International Max-Planck Research School for Intelligent Systems (IMPRS-IS) for the support.}
\thanks{Code: \url{https://github.com/Zwoelf12/compare_croco_scvx_komo_casadi}}%
}

\begin{document}

\maketitle
\thispagestyle{empty}
\pagestyle{empty}

\begin{abstract}
Quadrotors are agile flying robots that are challenging to control. Considering the full dynamics of quadrotors during motion planning is crucial to achieving good solution quality and small tracking errors during flight. Optimization-based methods scale well with high-dimensional state spaces and can handle dynamic constraints directly, therefore they are often used in these scenarios. The resulting optimization problem is notoriously difficult to solve due to its nonconvex constraints. In this work, we present an analysis of four solvers for nonlinear trajectory optimization (KOMO, direct collocation with SCvx, direct collocation with CasADi, Crocoddyl) and evaluate their performance in scenarios where the solvers are tasked to find minimum-effort solutions to geometrically complex problems and problems requiring highly dynamic solutions. Benchmarking these methods helps to determine the best algorithm structures for these kinds of problems.
\end{abstract}

\section{Introduction}

In recent years, multirotors have risen in popularity in academia and industry due to their exceptional agility and simple mechanical design. However, motion planning for these under-actuated systems requires to consider the dynamical constraints and is computationally expensive. While sampling- and search-based approaches to motion planning have strong theoretical guarantees regarding completeness, they suffer from the curse of dimensionality and scale exponentially with the dimension of the state space. Optimization-based methods on the other hand, often provide speed advantages in high dimensional state spaces (only polynomial scaling with the number of state space dimensions) and better quality of the found solutions. Since the motion planning problem is notoriously difficult to solve, approximate solutions are frequently applied. Common approximations are using the double integrator model of a point mass \cite{Augugliaro2012a} or the differential-flatness property of the quadrotor model \cite{Mellinger2011}, which allows to compute and follow splines as trajectories. Unfortunately, such approaches neglect the real model dynamic completely or cannot take the limited motor forces of a real quadrotor into account and produce conservative motions. In this paper we consider the full nonlinear dynamic model of the system to formulate a minimum-effort control problem and benchmark four different optimization-based methods on four scenarios, including geometrically challenging obstacle avoidance and highly dynamic maneuvers. %

\FloatBarrier

\section{Approach}

\subsection{Quadrotor Model}

We consider a quadrotor model with state vector $x = [p, v,q,\omega_B]^T \in \mathbb{R}^{13}$, where $p \in \mathbb{R}^{3}$ is the position, $v \in \mathbb{R}^{3}$ is the velocity (both in the inertial frame), $q \in \mathbb{H}$ is the unit quaternion rotation (parametrizing the rotation matrix $\mathbf{R}(q) \in SO(3)$), and $\omega_B \in \mathbb{R}^{3}$ is the rotational velocity in the body frame. The dynamic model is derived from the Newton-Euler equations for rigid bodies with 6 degrees of freedom (DoF) \cite{Sun2021}:

\begin{center}
  \begin{minipage}[b]{.2\textwidth}
  \begin{equation}
      \dot{p} = v 
      \label{eq:p_dot}
  \end{equation}
    \smallskip
  \begin{equation}
      \dot{v} = \frac{1}{m}\mathbf{R}(q)f_T + g
      \label{eq:v_dot}
  \end{equation}
  \end{minipage}
  \quad
  \begin{minipage}[b]{.255\textwidth}
    \begin{equation}
      \dot{q} = \frac{1}{2}q \otimes 
      \left(\begin{array}{c}
      0 \\
      \omega_B  \\
      \end{array}\right)
      \label{eq:q_dot}
    \end{equation}
      \smallskip
    \begin{equation}
      \dot{\omega}_B = \mathbf{J}^{-1}(\tau - \omega_B \times \mathbf{J} \omega_B).
      \label{eq:om_dot}
    \end{equation}
  \end{minipage}
\end{center}

Here, $\mathbf{J}$ denotes the inertia matrix of the  multirotor (in body frame), $m$ its mass, $g$ the gravity vector, $f_T$ the applied combined thrust, $\tau$ the applied torque, and $\otimes$ denotes the quaternion product. 

The total thrust vector $f_T$ and the acting torque $\tau$ result from the multirotor geometry and the acting forces generated by the propellers as follows:

\begin{align*}
f_T = 
      \left(\begin{array}{c}
      0 \\
      0  \\
      \sum_i f_i
      \end{array}\right), &
\tau =
      \left(\begin{array}{c}
      \frac{\sqrt{2}l}{2}(-f_1-f_2+f_3+f_4) \\
      \frac{\sqrt{2}l}{2}(-f_1+f_2+f_3-f_4) \\
      \kappa_{\tau}(f_1 - f_2 + f_3 - f_4) \\
      \end{array}\right),
\end{align*}

where $\kappa_{\tau}$ is the torque constant and $l$ is the arm length of the multirotor.
The forces $f_i$ are related to the controllable rotor speed $\omega_{i}$ by the thrust coefficient $\kappa_f$ according to $f_i = \kappa_f \omega_{i} ^2$.
We use $u = [f_1,\ldots,f_4]^T$ as controls. 

\subsection{Non-linear Program (NLP) Formulation}

We formulate the following discrete-time optimization problem over $N$ steps.

\begin{alignat}{2}
    &\min_{\hat{x},\hat{u}}          & \qquad & \sum_{k=0}^{N-1}\|u_k\|^2, \label{eq:opt:obj}\\
    &\text{subject to} &        &  x_{k+1} = step(x_k,u_k), \label{eq:opt:step}\\
    &                  &        &  S_k \cap O_i = \emptyset , \forall i\in\{1,...,n_{obs}\} ,\label{eq:opt:obstacles} \\
    &                  &        &  x_0 = x_s, x_N = x_f, \label{eq:opt:start_goal}\\
    &                  &        &   x_{k_{m,i}} = x_{m,i} \forall i \in \{1,...,n_{is}\}, \label{eq:opt:intermed}\\ 
    &                  &        &  x_k \leq x_{\max}, x_k \geq x_{\min},\label{eq:opt:statespace}\\
    &                  &        &  u_k \leq u_{\max}, u_k \geq u_{\min}.\label{eq:opt:actionspace}
\end{alignat}

With states $\hat{x} = (x_0,x_1,...,x_N), x_k \in \mathbb{R}^{13}$ and controls $\hat{u} = (u_0,u_1,...,u_{N-1}), u_k \in \mathbb{R}^4$.
Here, \eqref{eq:opt:step} captures the dynamics according to \eqref{eq:p_dot} -- \eqref{eq:om_dot}; \eqref{eq:opt:obstacles} avoid collisions (the multirotor, approximated by the sphere $S_k$, does not intersect with the obstacles $O_i~\in~\mathcal{O}$, $ i~\in~\{1,...,n_{obs}\}$); \eqref{eq:opt:start_goal} enforces that the trajectory starts in an initial state $x_s$ and ends in a final state $x_f$; \eqref{eq:opt:intermed} enforces intermediate states $x_{m,i}$ with $i \in \{1,...,n_{is}\}$; \eqref{eq:opt:statespace} limits the states to be within user-specified bounds; \eqref{eq:opt:actionspace} limits the motor force magnitudes (for the highly dynamic scenarios, the optimal manouvers require reaching the limits); and \eqref{eq:opt:obj} minimizes the required force as an approximation of the used energy.

\subsection{Used Methods}

The discrete problem is implemented in four different trajectory optimization frameworks. These use different transcriptions and algorithms for solving the nonlinear problem. All methods discretize the continuous-time problem at $N=100$ time points.

\subsubsection{Direct Collocation (DC)}

Two of the compared frameworks use direct collocation.

\textit{Sequential Convex Programming (SCP):}
 SCP methods approximate the non-convex discrete optimization problem iteratively as a convex sub-problem updating the approximation according to newly obtained sub-problem solutions. Advantages of SCP methods include that they have meaningful theoretical guarantees regarding algorithmic complexity and performance~\cite{Malyuta2021} and that they are agnostic to the choice of the convex solver. As SCP method we use SCvx~\cite{Mao2016}, which we implement in Python following \cite{Malyuta2021}. Positions, orientations, and linear and angular velocities as well as motor forces are introduced as optimization variables and the continuous dynamics are discretized using an explicit Euler integration scheme ($step$-function). The collision avoidance constraints are formulated using signed distances calculated with the flexible collision library (fcl)~\cite{Pan2012}. We formulate the discrete and convex sub-problem in every iteration using CVXPY~\cite{Diamond16} and solve it with ECOS~\cite{Domahidi2015}. ECOS is a solver specialized in solving convex problems, is written in C, uses a log-barrier method to formulate a series of unconstrained problems, and employs Newton's method as an inner loop.

\textit{CasADi + IPOPT:}
Additionally, we solve the NLP directly with a nonlinear solver. We implement the discrete NLP using CasADi \cite{Andersson18}. CasADi is an open-source software framework for numerical optimization and can be used to formulate NLPs like optimal control problems. To formulate the problem, the positions, orientations, linear and angular velocities as well as the motor forces are introduced as decision variables. The continuous dynamics are discretized using a 4th-order Runge-Kutta integration scheme ($step$-function). To incorporate collision avoidance constraints, spherical keep-out zones were introduced. To solve the formulated problem CasADi internally uses IPOPT \cite{Waechter05} which is implemented in Fortran and C. IPOPT employs a primal-dual barrier method to formulate a sequence of unconstrained optimization problems which are solved using a damped Newton's method.

\subsubsection{K-Order Markov Optimization (KOMO)}
KOMO is a method for efficiently solving robot motion planning problems originally introduced in 2014 \cite{KOMO}. In comparison to other motion planning methods, KOMO represents the trajectory only in configuration space $(p, q)$ instead of $(p,v,q,\omega)$, computing differential quantities by finite differences of consecutive configurations. Optimization variables are therefore only the positions and orientations of the quadcopter and the motor forces. Due to the structure introduced by the short-term dependency of the Markov property of the trajectory optimization problem, the Jacobian and the pseudo-Hessian of the problem result in banded and banded-symmetric matrices which are efficient to compute, store, and factorize. The resulting dependency between the states at consecutive time points and forces is that of an implicit Euler integration scheme ($step$-function). Similar to SCvx, the formulated collision avoidance constraints are based on signed distances which are calculated using fcl. The formulated constrained NLP is solved using Augmented Lagrangian which internally uses Newton's method with line search to solve the unconstrained optimization problem in each iteration. The problem was implemented using the Python bindings of the RAI-Framework\footnote{\url{https://github.com/MarcToussaint/rai-python}}, which is implemented in C++.

\subsubsection{Differential Dynamic Programming (DDP)}
DDP is an algorithm for continuous optimal control based on Bellman’s principle of optimality. In each iteration, it uses a backward pass (in time) to build a local quadratic approximation of the cost-to-go value, and a forward pass to update the state and control trajectory. DDP directly accounts for the cost and the dynamic constraints in the forward-backward pass. The dynamics are discretized using an explicit Euler integration scheme ($step$-function). The goal, collision and state constraints are added using a squared penalty method. In contrast to the previously mentioned methods, DDP introduces only motor forces as decision variables, optimizing the state trajectory implicitly. The collision avoidance constraints are formulated using signed distances and fcl as well. Specifically, we use BOX-FDDP \cite{Mastalli2020, Mastalli2021}, which considers control limits directly in the forward-backward pass (instead of using a squared penalty) and can be efficiently warm-started using an infeasible initial guess, offering better globalization capabilities. The main advantage of DDP is that the Markov structure of the trajectory optimization problem (like KOMO) can directly be exploited to compute second-order search directions, which is considerably faster than using general-purpose matrix factorization techniques with sparse matrices. Moreover, when the dynamical constraints are highly non-linear, the DDP recursion is often more efficient than adding these constraints in constrained optimization methods (e.g., Augmented Lagrangian). In our benchmark, we use Crocoddyl \cite{Mastalli2020}, an open-source DDP solver that provides an efficient C++ implementation (e.g., no dynamic memory allocation, fast linear algebra operations using Eigen). 

\section{Experimental Results}

We evaluate the four different solvers in four scenarios. All scenarios were solved 30 times on a workstation (AMD Ryzen Threadripper PRO 5975WX @ 3.6 GHz, 64 GB RAM, Ubuntu 22.04), each with a different initial guess.
 
\subsection{Evaluated Scenarios}

\begin{figure}[t] 
    \centering
  \subfloat[Scenario 1\label{fig:Scenario1}]{%
       \includegraphics[trim = 0.7cm 0.cm 0.1cm 0.cm , clip, width=0.45\linewidth]{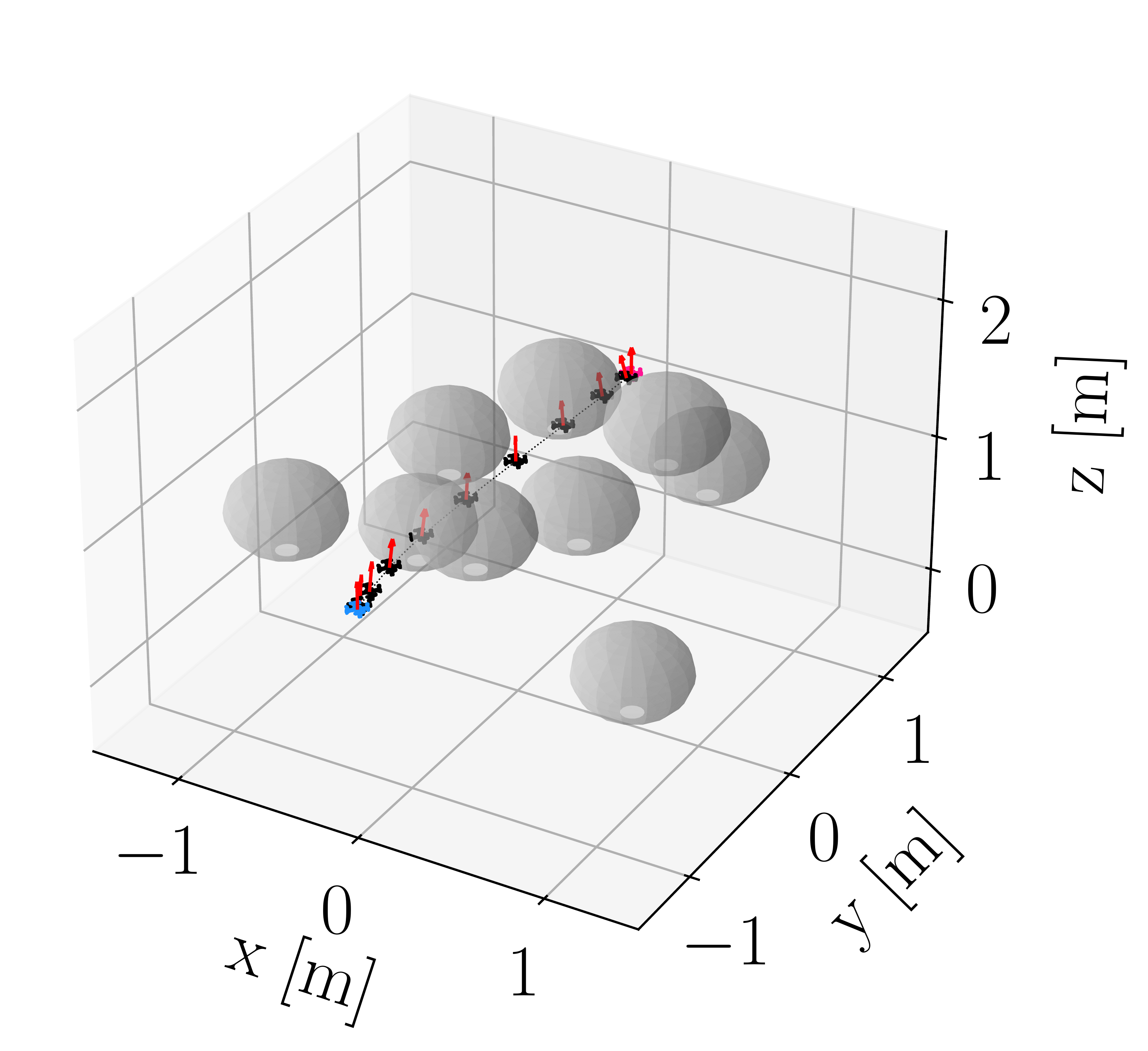}}
    \hfill
  \subfloat[Scenario 2\label{fig:Scenario2}]{%
        \includegraphics[trim = 0.7cm 0.cm 0.1cm 0.cm , clip, width=0.45\linewidth]{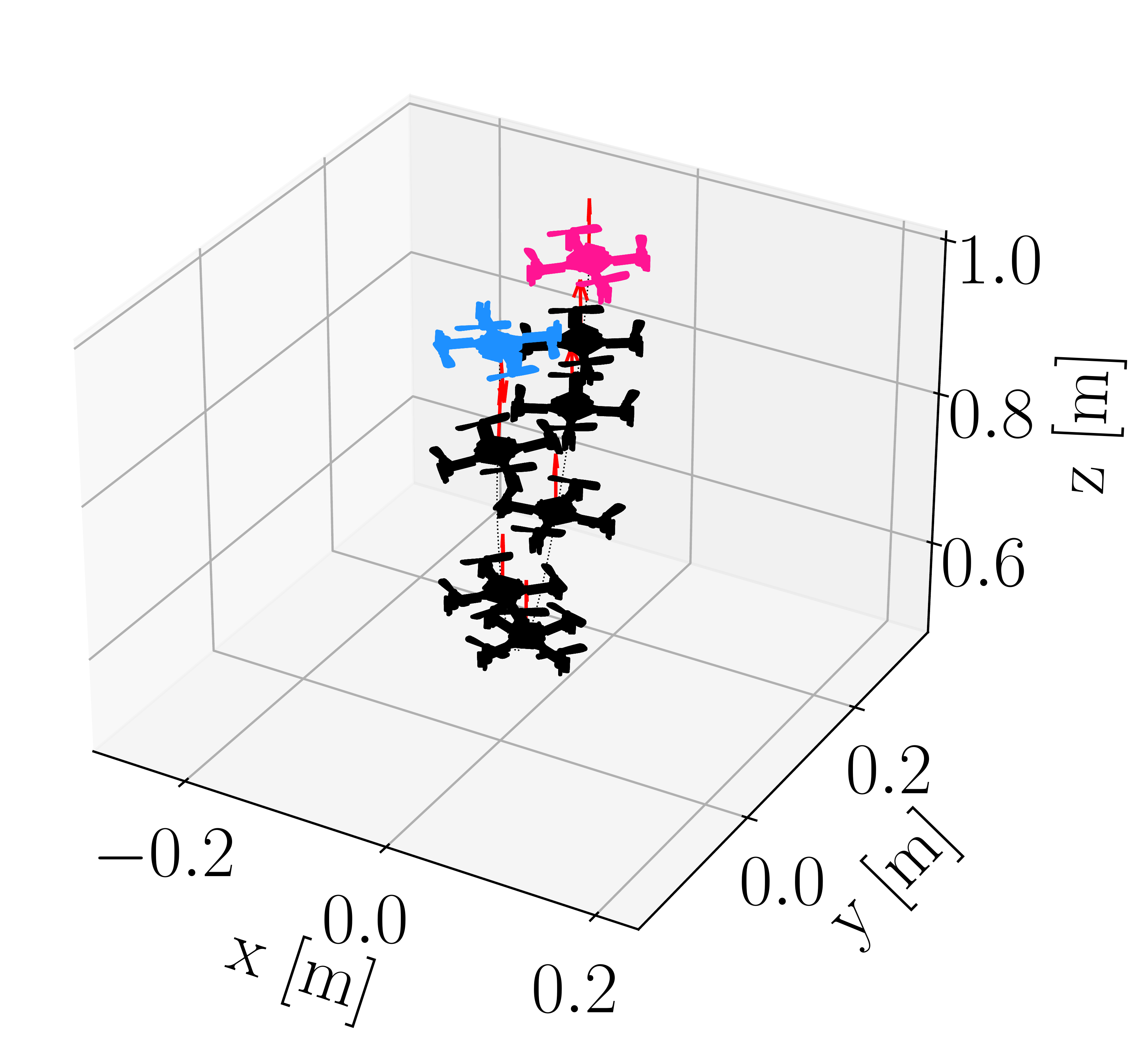}}
    \hfill
  \subfloat[Scenario 3\label{fig:Scenario3}]{%
        \includegraphics[trim = 0.7cm 0.cm 0.1cm 0.cm , clip, width=0.45\linewidth]{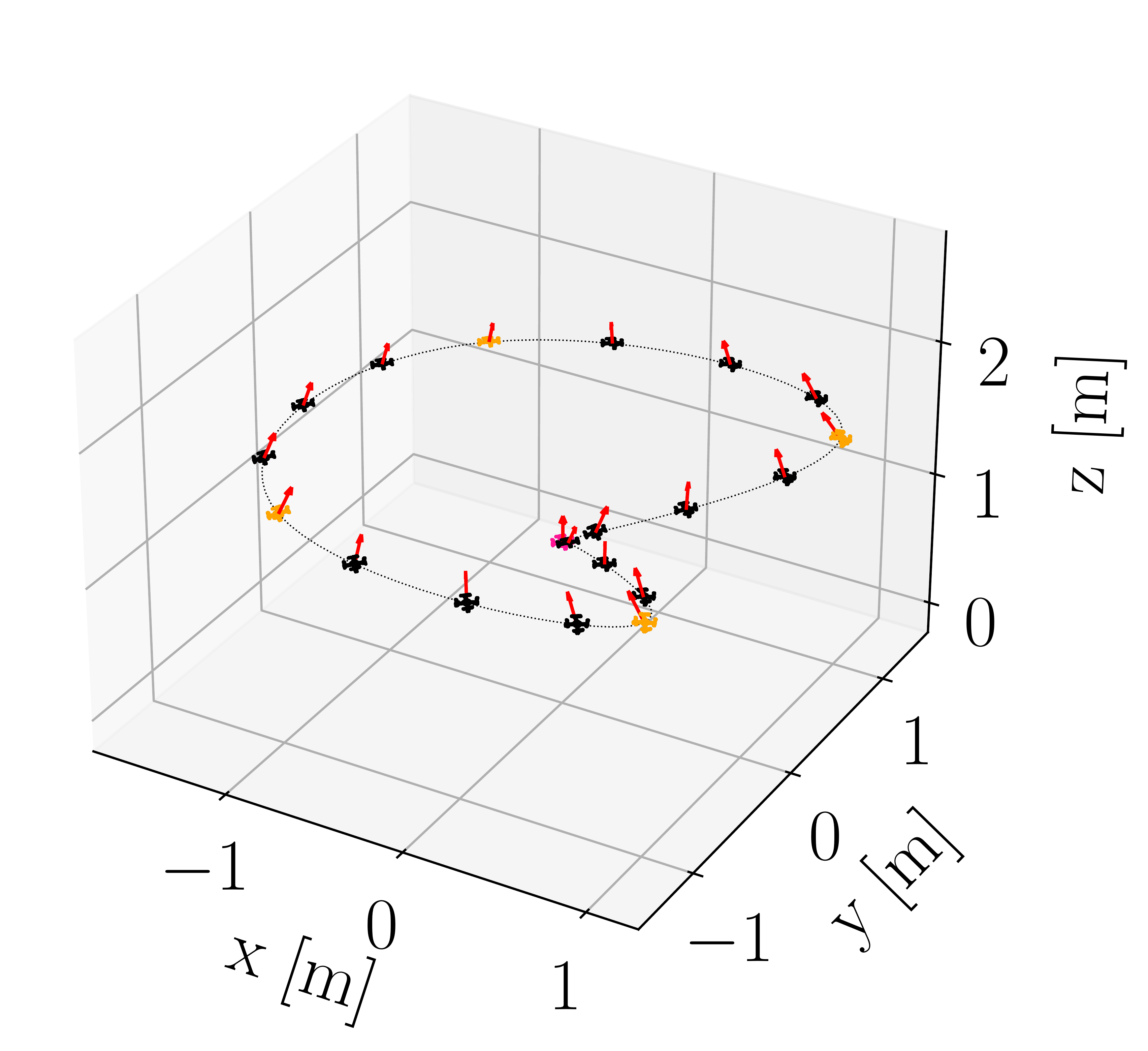}}
    \hfill
  \subfloat[Scenario 4\label{fig:Scenario4}]{%
        \includegraphics[trim = 0.6cm 0.cm 0.1cm 0.cm , clip, width=0.45\linewidth]{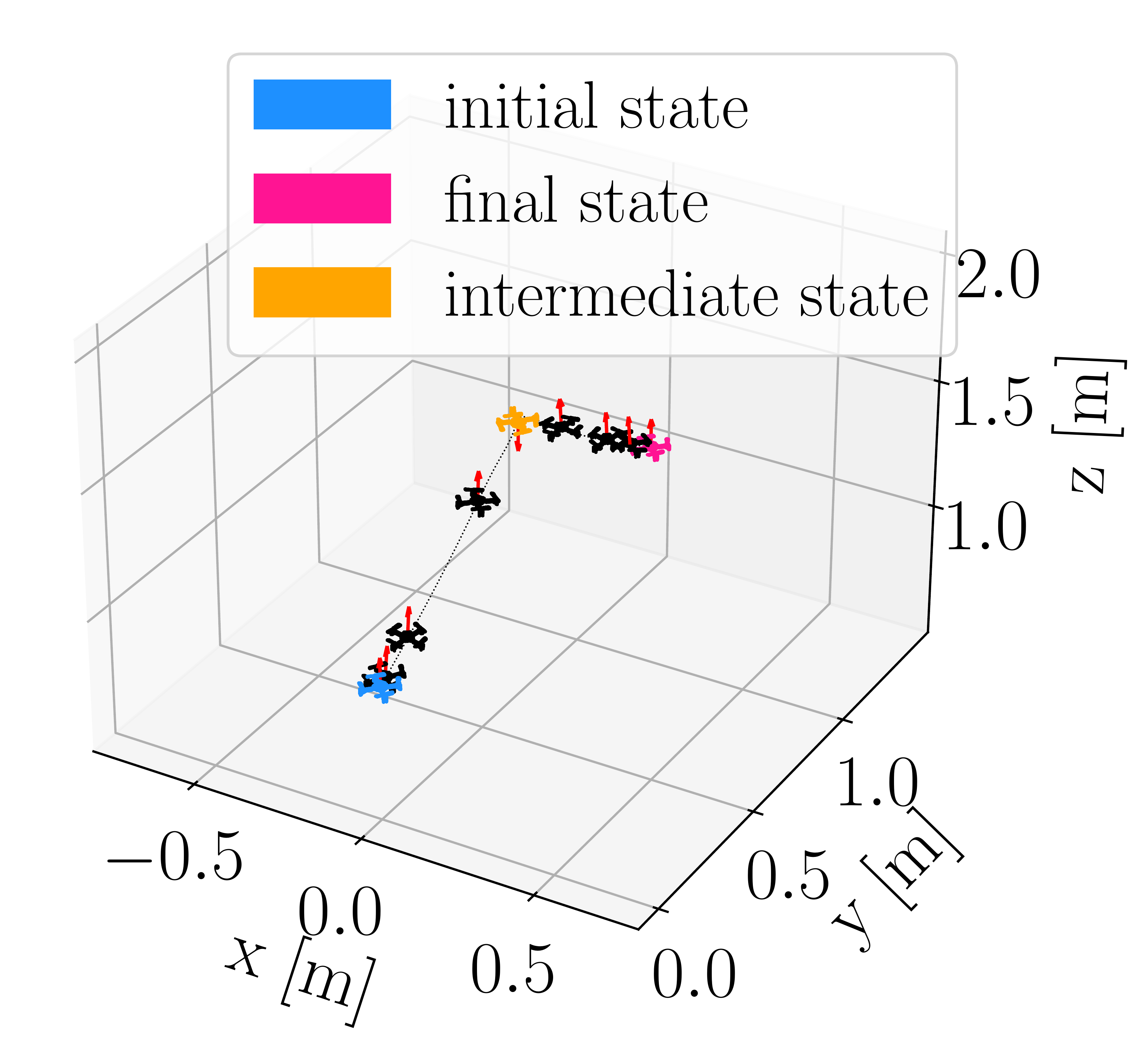}}
  \caption{Example trajectories for the chosen scenarios. The black quadrotors represent a possible trajectory solving the problem and the red arrows indicate the $z$-axis of the model.}
  \label{fig:Probs} 
\end{figure}

The selected scenarios are shown in \cref{fig:Probs} and include geometrically complex problems and problems requiring highly dynamic solutions. Here the red arrows indicate the $z$-axis of the body frame. In scenario 1, a trajectory has to be found leading through an environment cluttered with spherical obstacles. For scenario 2, the quadcopter has to recover from an upside-down position. In scenario 3, the optimizers have to find a trajectory following 4 waypoints. The intermediate constraints in scenario 3 restrict only the position of the quadcopter. For scenario 4 an intermediate constraint is introduced forcing the orientation to be upside-down in the middle of the trajectory.

\subsection{Initial Guess}
For scenarios 1 and 3, the initial guess was calculated by linear interpolation between the initial, intermediate and final states for positions. For the orientations, the initial guess was obtained by spherical linear interpolation and the motor forces were initialized such that the gravitational force was compensated (hover condition). The rotational and linear velocities were initialized with zero. Gaussian noise related to the scaled maximum range of each quantity was added to all states and forces. Note that all solvers received the same initial guess. For scenarios 2 and 4 the solvers were initialized with the initial position and the hovering orientation. The forces and the linear and rotational velocities were initialized in the same fashion as for scenarios 1 and 3 and the noise was added following the same pattern as well.

\subsection{Results}
\begin{figure}[t] 
    \centering
  \subfloat[Scenario 1\label{fig:CostScenario1}]{%
       \includegraphics[trim = 0.8cm 0.8cm 0.8cm 0.8cm , clip, width=0.49\linewidth]{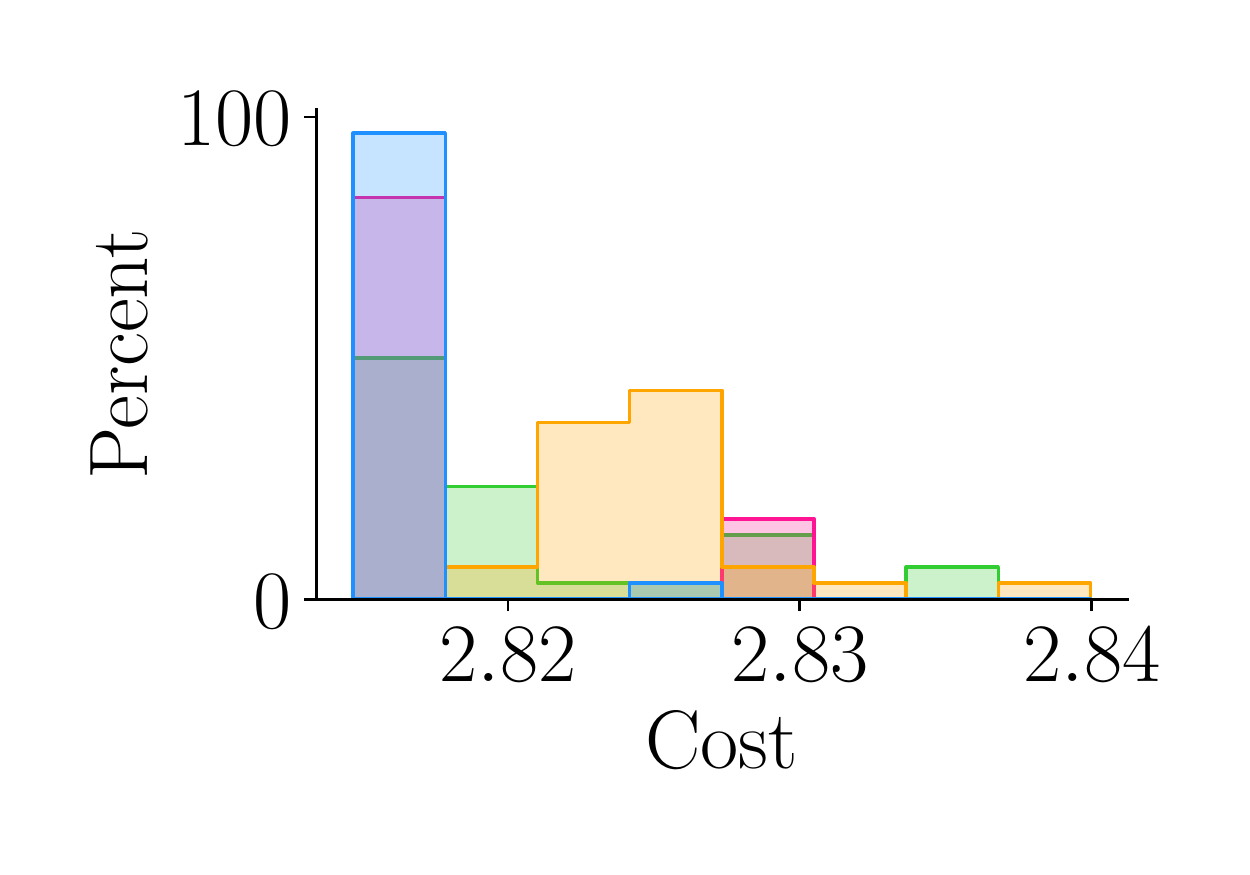}}
    \hfill
  \subfloat[Scenario 2\label{fig:CostScenario2}]{%
        \includegraphics[trim = 0.8cm 0.8cm 0.8cm 0.8cm  , clip, width=0.49\linewidth]{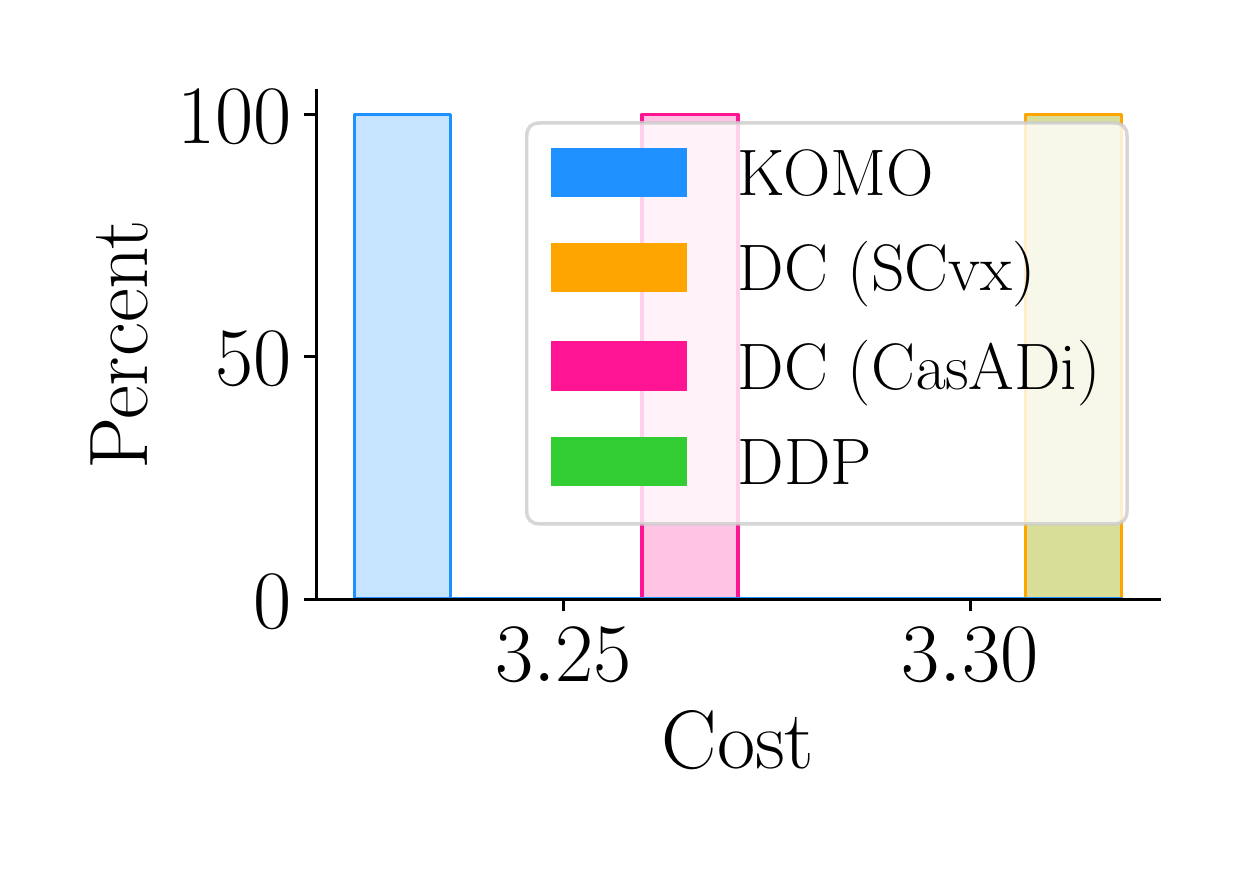}}
    \hfill
  \subfloat[Scenario 3\label{fig:CostScenario3}]{%
        \includegraphics[trim = 0.8cm 0.8cm 0.8cm 0.8cm , clip, width=0.49\linewidth]{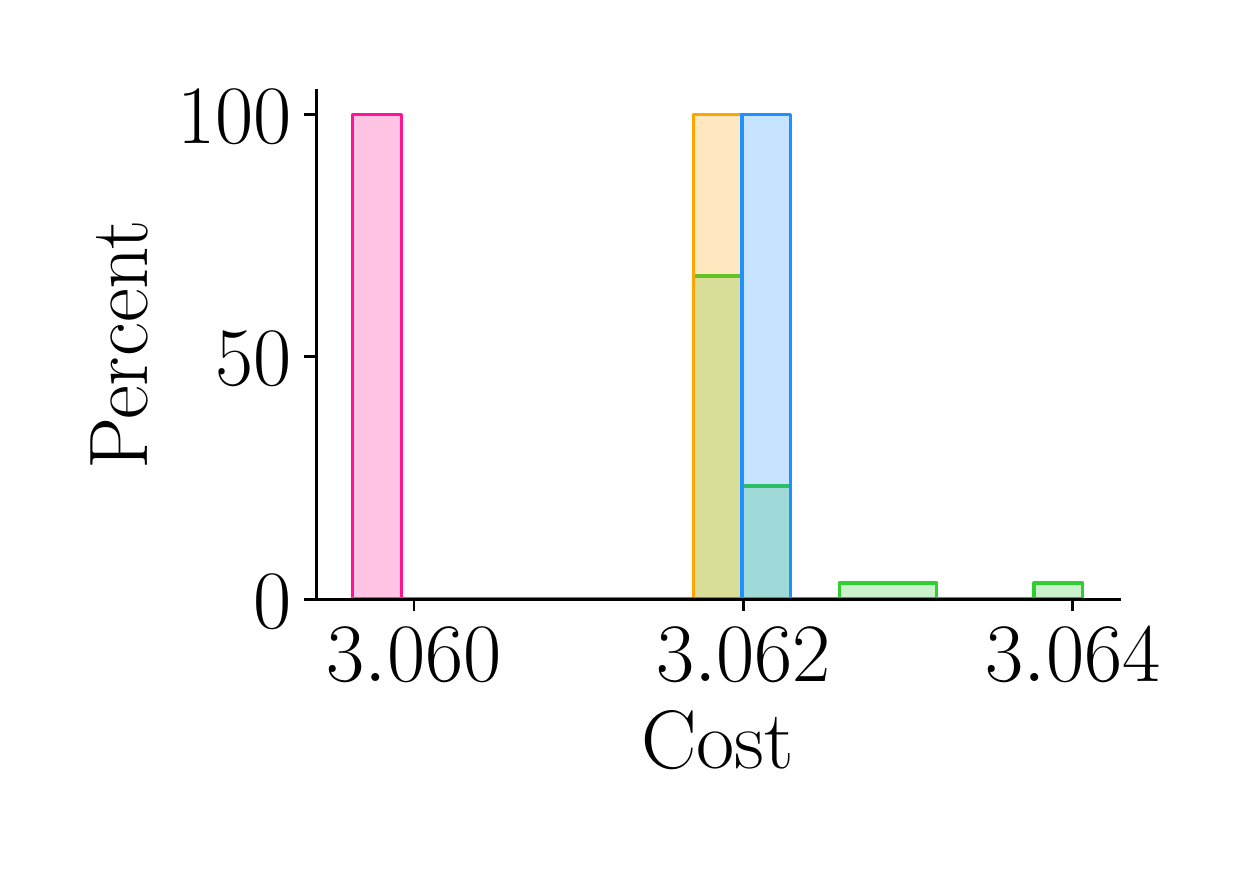}}
    \hfill
  \subfloat[Scenario 4\label{fig:CostScenario4}]{%
        \includegraphics[trim = 0.8cm 0.8cm 0.8cm 0.8cm , clip, width=0.49\linewidth]{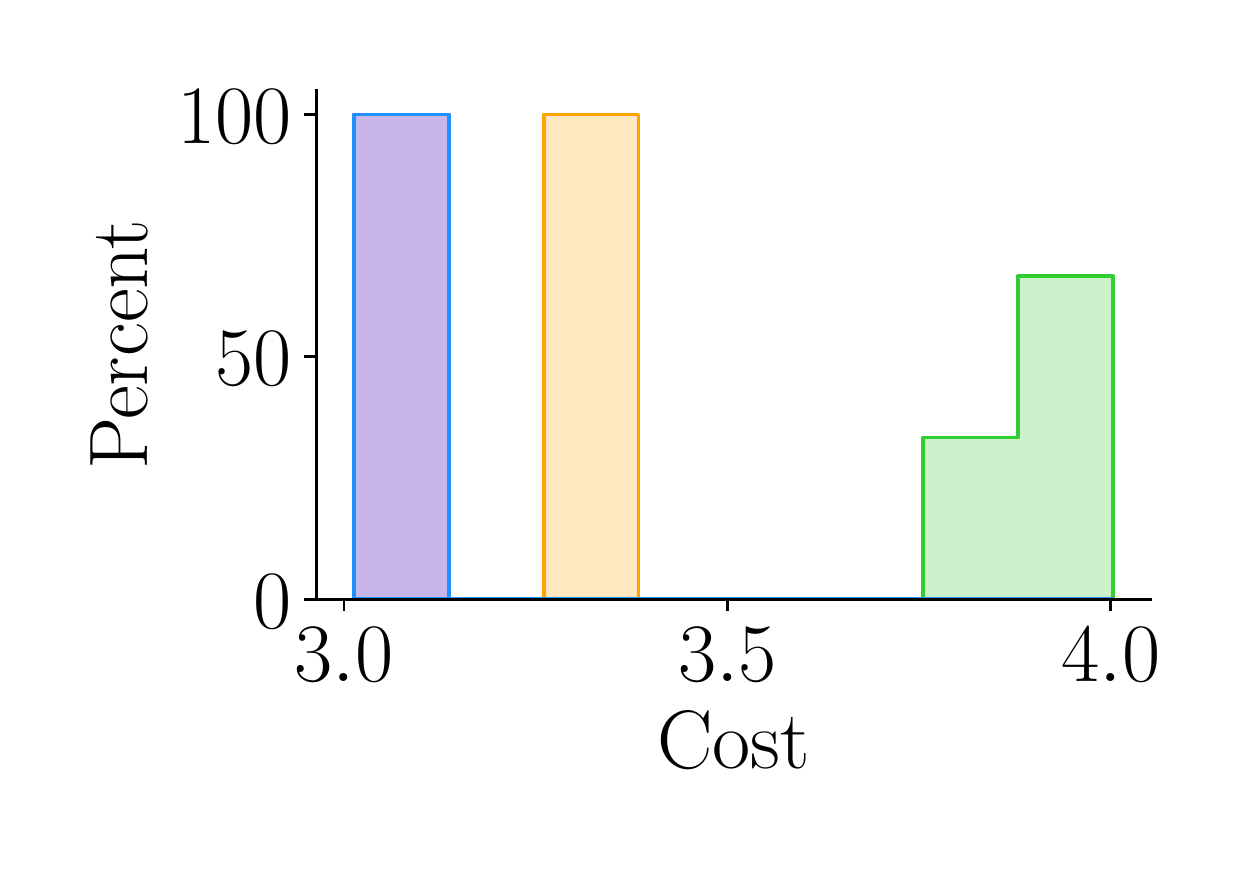}}
  \caption{Histograms of the optimal values of each solver for all four scenarios.}
  \label{fig:costs} 
\end{figure}

Throughout the experiments, all solvers found a feasible solution in all 30 runs. To evaluate the solver's solution quality, we report the converged values of the objective function. \Cref{fig:costs} shows the histograms of the objective function values. For the geometrically challenging scenarios (1,3), all solvers converge to similar optimal solutions, with a difference between the objective values of less than 1\%. For the highly dynamic scenarios (2,4), the solvers converge to different optimal values, with KOMO having in general the lowest optimal value and DDP the highest. The distributions of the optimal values for each solver in this scenario are narrow, meaning that the solvers converge to the same optimal value regardless of the noise in the initial estimate. 

\begin{figure}[t] 
    \centering
  \subfloat[Scenario 1\label{fig:IterScenario1}]{%
       \includegraphics[trim = 0.8cm 0.8cm 0.8cm 0.8cm , clip, width=0.49\linewidth]{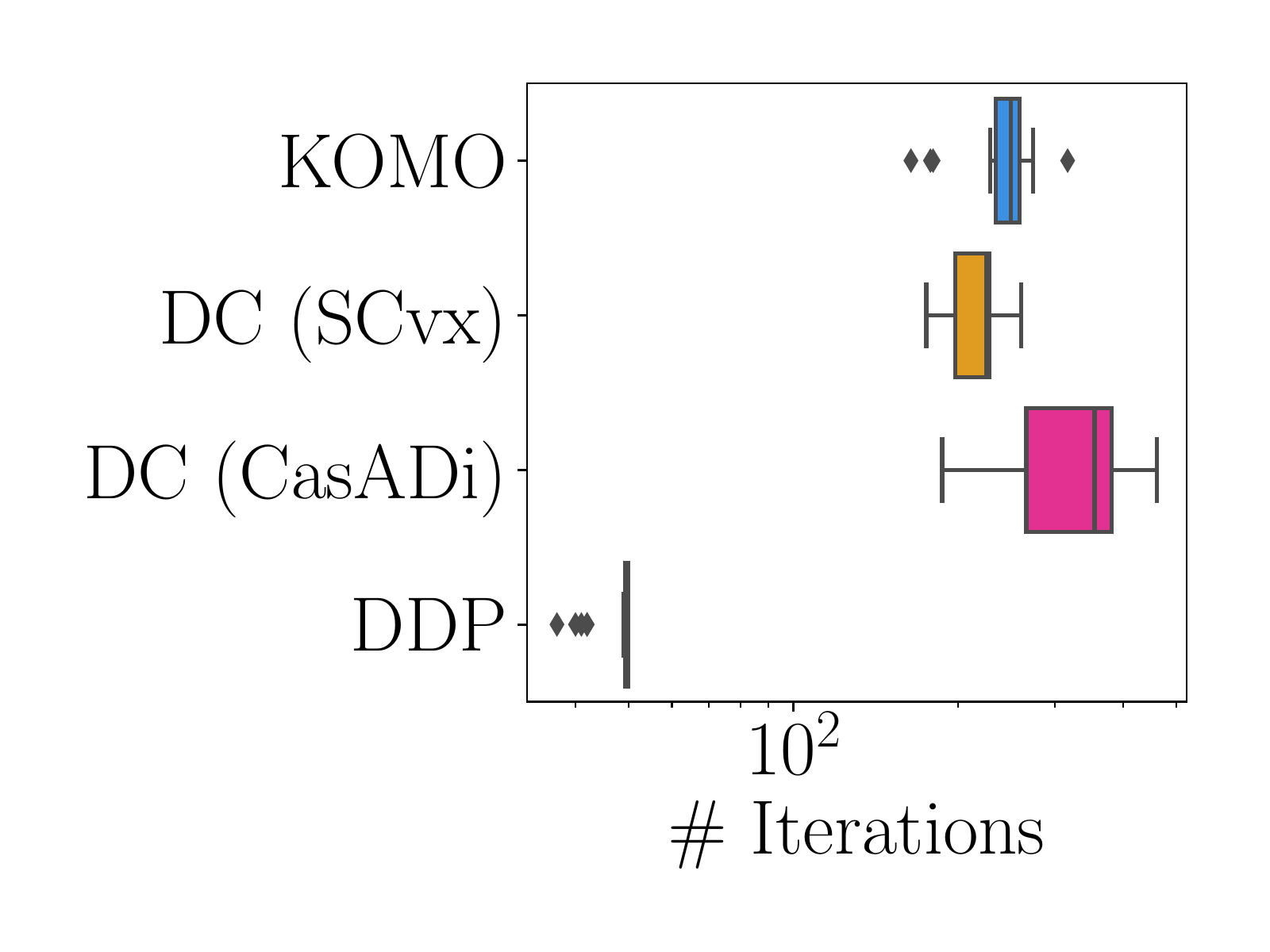}}
    \hfill
  \subfloat[Scenario 2\label{fig:IterScenario2}]{%
        \includegraphics[trim = 0.8cm 0.8cm 0.8cm 0.8cm , clip, width=0.49\linewidth]{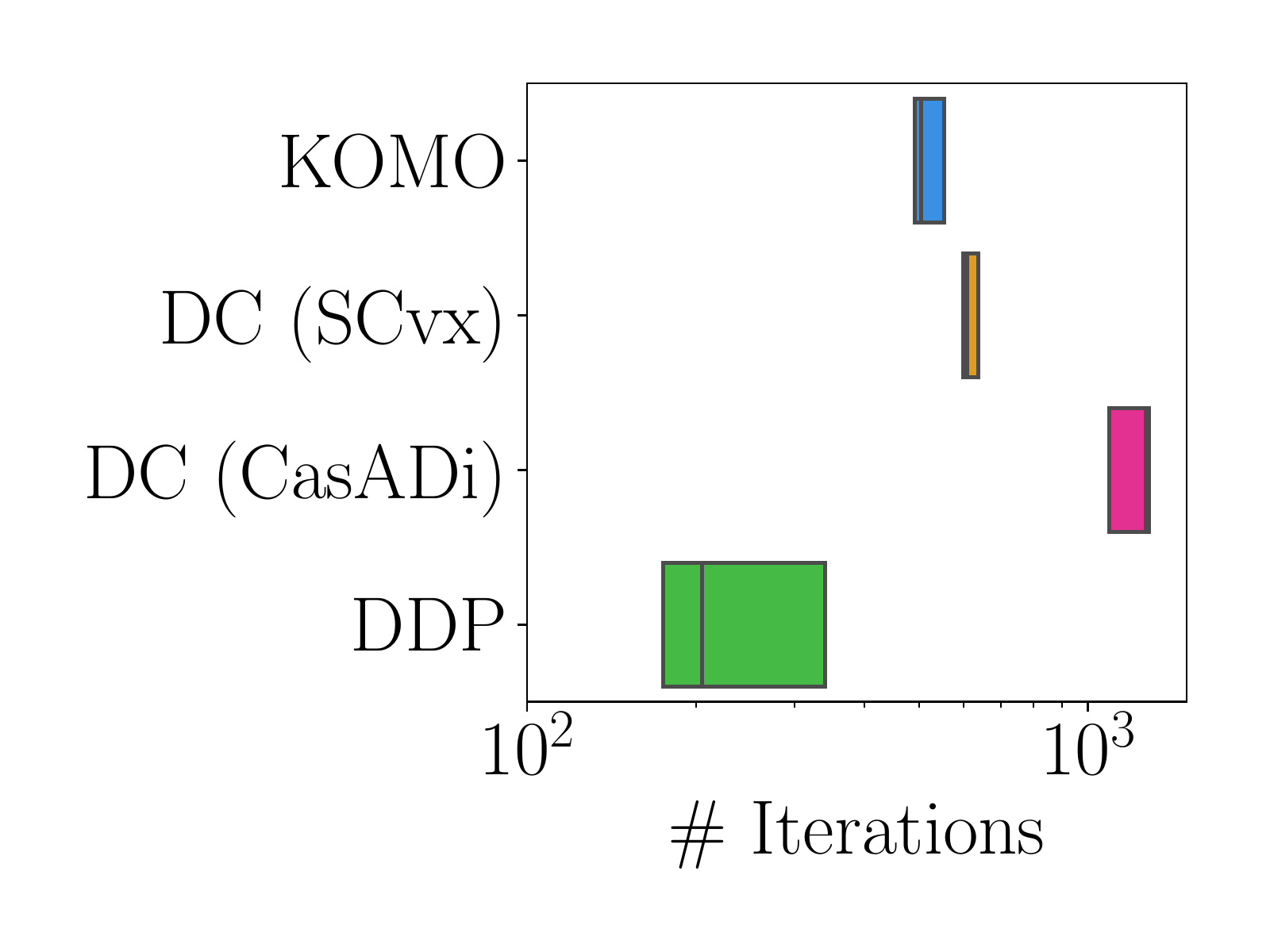}}
    \hfill
  \subfloat[Scenario 3\label{fig:IterScenario3}]{%
        \includegraphics[trim = 0.8cm 0.8cm 0.8cm 0.8cm , clip, width=0.49\linewidth]{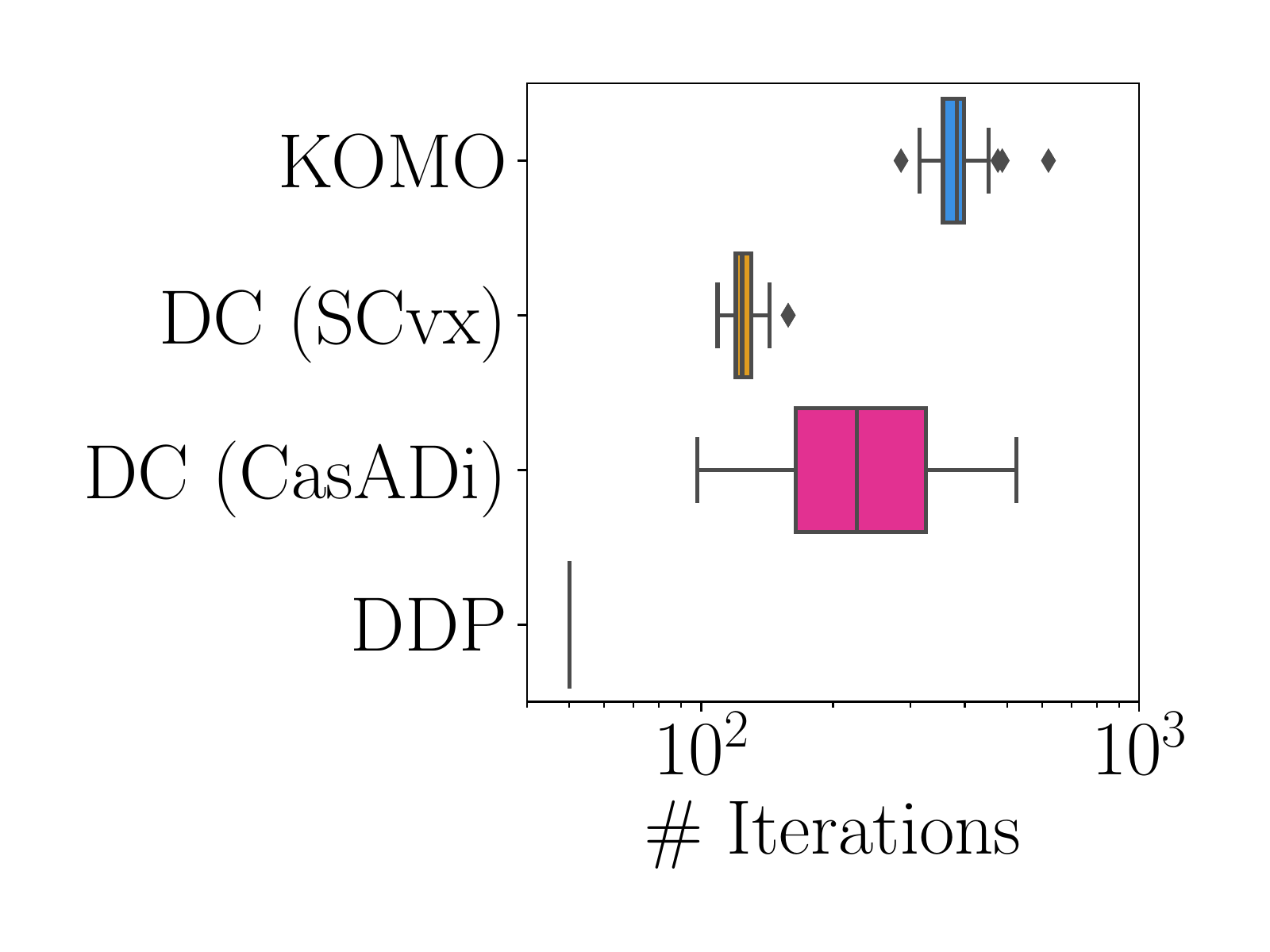}}
    \hfill
  \subfloat[Scenario 4\label{fig:IterScenario4}]{%
        \includegraphics[trim = 0.8cm 0.8cm 0.8cm 0.8cm , clip, width=0.49\linewidth]{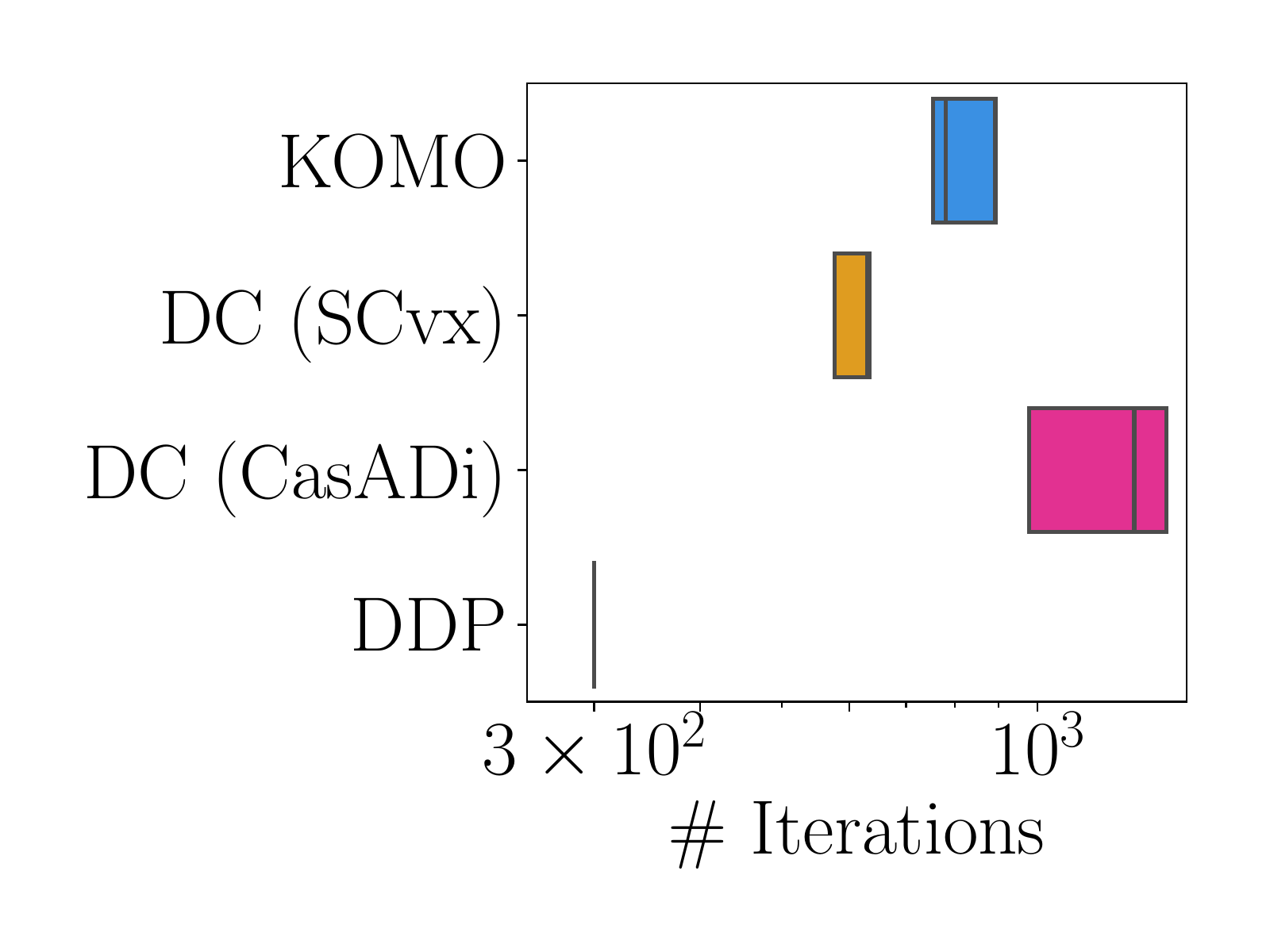}}
  \caption{Box-plots of the number of Newton iterations for KOMO, DC with SCvx and DC with CasADi and the number of DDP iterations.}
  \label{fig:iterations} 
\end{figure}

To evaluate the effort required to solve the problems, the number of Newton method iterations for DC with SCvx, KOMO, and DC with CasADi and the number of DDP iterations are given in \cref{fig:iterations}. For algorithms using Newton's method, the number of Newton iterations is a reasonable measure of computational effort, since computing the step direction by solving a linear system of equations is the most computationally expensive operation. Note that a DDP iteration is approximately equivalent to a newton iteration regarding computational effort. Throughout the experiments, DDP required an order of magnitude fewer numbers of iterations to converge to an optimum in all runs compared to the other solvers. For scenarios 3 and 4, DDP terminates due to an iteration limit in every run. But even with more iterations, the final objective value does not change. In the highly dynamic scenarios (2,4), direct collocation with CasADi required the most Newton iterations to converge in all runs. In general, it can be observed that the variance of the number of required iterations is highest for DC with CasADi.

In addition, we evaluated the time required in the optimizer to solve the NLP. The results are shown in \cref{tab:times}. In all scenarios, DDP takes the least time by one to two orders of magnitude and solves all problems in less than one second. DC with CasADi takes the most time in all scenarios. In highly dynamic scenarios (2,4), DC with CasADi takes almost an order of magnitude longer than DC with SCvx and KOMO. Note that the comparison regarding runtime does not indicate algorithmic advantages, since the solvers are partly written in different programming languages.

\begin{table}
    \centering
    \caption{The mean and the standard deviation (gray subscript) of the time in seconds spent in the optimizer for the different solvers and scenarios. The best results are bold.}
    \begin{tabular}{|c|r|r|r|r|}
        \hline
        \textbf{Scenario} & \textbf{KOMO} & \textbf{DC (SCvx)} & \textbf{DC (CasADi)} & \textbf{DDP} \\
        \hline
        \hline
        \textbf{1} & $7.8_{\;\textcolor{gray}{0.9}}$  & $29.9_{\;\textcolor{gray}{2.3}}$ & $66.0_{\;\textcolor{gray}{15.7}}$ & $\mathbf{0.20_{\;\textcolor{gray}{0.03}}}$ \\ 
        \hline
        \textbf{2} & $17.4_{\;\textcolor{gray}{1.0}}$  & $23.1_{\;\textcolor{gray}{0.3}}$ & $259.8_{\;\textcolor{gray}{12.2}}$ & $\mathbf{0.93_{\;\textcolor{gray}{0.30}}}$ \\ 
        \hline
        \textbf{3} & $11.3_{\;\textcolor{gray}{1.7}}$  & $9.7_{\;\textcolor{gray}{0.3}}$ & $48.6_{\;\textcolor{gray}{20.0}}$ & $\mathbf{0.04_{\;\textcolor{gray}{0.00}}}$ \\ 
        \hline
        \textbf{4} & $27.0_{\;\textcolor{gray}{4.5}}$  & $18.9_{\;\textcolor{gray}{0.3}}$ & $237.9_{\;\textcolor{gray}{40.8}}$ & $\mathbf{0.93_{\;\textcolor{gray}{0.08}}}$ \\ 
        \hline
    \end{tabular}
    \label{tab:times}
\end{table}

\section{Conclusion and Future Work}
To objectively compare and evaluate different trajectory optimization techniques we need standard benchmarks.
To this end, we benchmark four different optimization-based solvers on dynamically and geometrically challenging scenarios of multirotor flight.
We tune each solver, since the performance highly depends on the choice of user-defined weights and the parameters of the algorithms.
Our results show that KOMO achieves the lowest objective function values across all scenarios, while DDP requires the least amount of time and iterations.
We conclude that solvers which leverage the structure of trajectory optimization problems (KOMO and DDP) are beneficial over formulating the trajectory optimization problem as a standard NLP, as for DC with CasADi.

For the obtained optimal values, the cause for the consistent computation of the higher-cost solutions of DDP in Scenario 4 should be investigated. One conjecture is that this is related to small differences in the constraint formulation and slight violations of some constraints in the other solvers. 
For the computational effort, comparing the number of iterations can be misleading since the effort for DDP iterations and Newton iterations is not identical and we do not account for the number of line search iterations for Newton's method.
While we compare specific implementations, it remains an open issue whether to attribute the performance differences to a better algorithmic design or a better implementation (e.g., programming language, efficiency of the linear algebra operations).

\bibliography{bibliography}
\bibliographystyle{ieeetr}

\end{document}